\documentclass[10pt,twocolumn,letterpaper]{article}

\usepackage{wacv}
\usepackage{times}
\usepackage{epsfig}
\usepackage{graphicx}
\usepackage{amsmath}
\usepackage{amssymb}
\usepackage{booktabs}
\usepackage{multirow}
\usepackage{soul}

\DeclareMathAlphabet{\pazocal}{OMS}{zplm}{m}{n}

\def\wacvPaperID{2059}

\wacvalgorithmstrack

\wacvfinalcopy 

\ifwacvfinal
\usepackage[breaklinks=true,bookmarks=false]{hyperref}
\else
\usepackage[pagebackref=true,breaklinks=true,colorlinks,bookmarks=false]{hyperref}
\fi

\begin{document}

\title{ Hear The Flow: Optical Flow-Based Self-Supervised \\ Visual Sound Source Localization }

\author{Dennis Fedorishin\thanks{Equal contribution authors in alphabetic order} \and Deen Dayal Mohan\footnotemark[1] \and Bhavin Jawade \and Srirangaraj Setlur \and Venu Govindaraju \\
University at Buffalo, Buffalo, New York, USA \\
{\tt\small \{dcfedori,dmohan,bhavinja,setlur,govind\}@buffalo.edu}
}

\maketitle
\thispagestyle{empty}

\begin{abstract}

Learning to localize the sound source in videos without explicit annotations is a novel area of audio-visual research. Existing work in this area focuses on creating attention maps to capture the correlation between the two modalities to localize the source of the sound. In a video, oftentimes, the objects exhibiting movement are the ones generating the sound. In this work, we capture this characteristic by modeling the optical flow in a video as a prior to better aid in localizing the sound source. We further demonstrate that the addition of flow-based attention substantially improves visual sound source localization. Finally, we benchmark our method on standard sound source localization datasets and achieve state-of-the-art performance on the Soundnet Flickr and VGG Sound Source datasets. Code: \url{https://github.com/denfed/heartheflow}.
\end{abstract}

\vspace{-9px}

\section{Introduction}

In recent years, the field of audio-visual understanding has become a very active area of research. This can be attributed to the large amount of video data being produced as part of user-generated content on social media and other platforms. Recent methods in audio-visual understanding have leveraged popular deep learning techniques to solve challenging problems such as action recognition \cite{gao2020listen}, deepfake detection \cite{zhou2021joint}, and other tasks. Given a video, one such task in audio-visual understanding is to locate the object in the visual space that is generating the prominent audio content. When observing a natural scene, it is often trivial for a human to localize the region/object from which the sound originates. One of the main reasons for this is the binaural nature of the human hearing sense. However, the majority of audio-visual data in digital media is monaural, which complicates audio localization tasks. Furthermore, naturally occurring videos do not have explicit annotations of the location of the source of the audio in the image. This makes the task of training deep neural networks to understand audio-visual associations for localization a challenging task.

\begin{figure}
\centering
\includegraphics[scale=0.062]{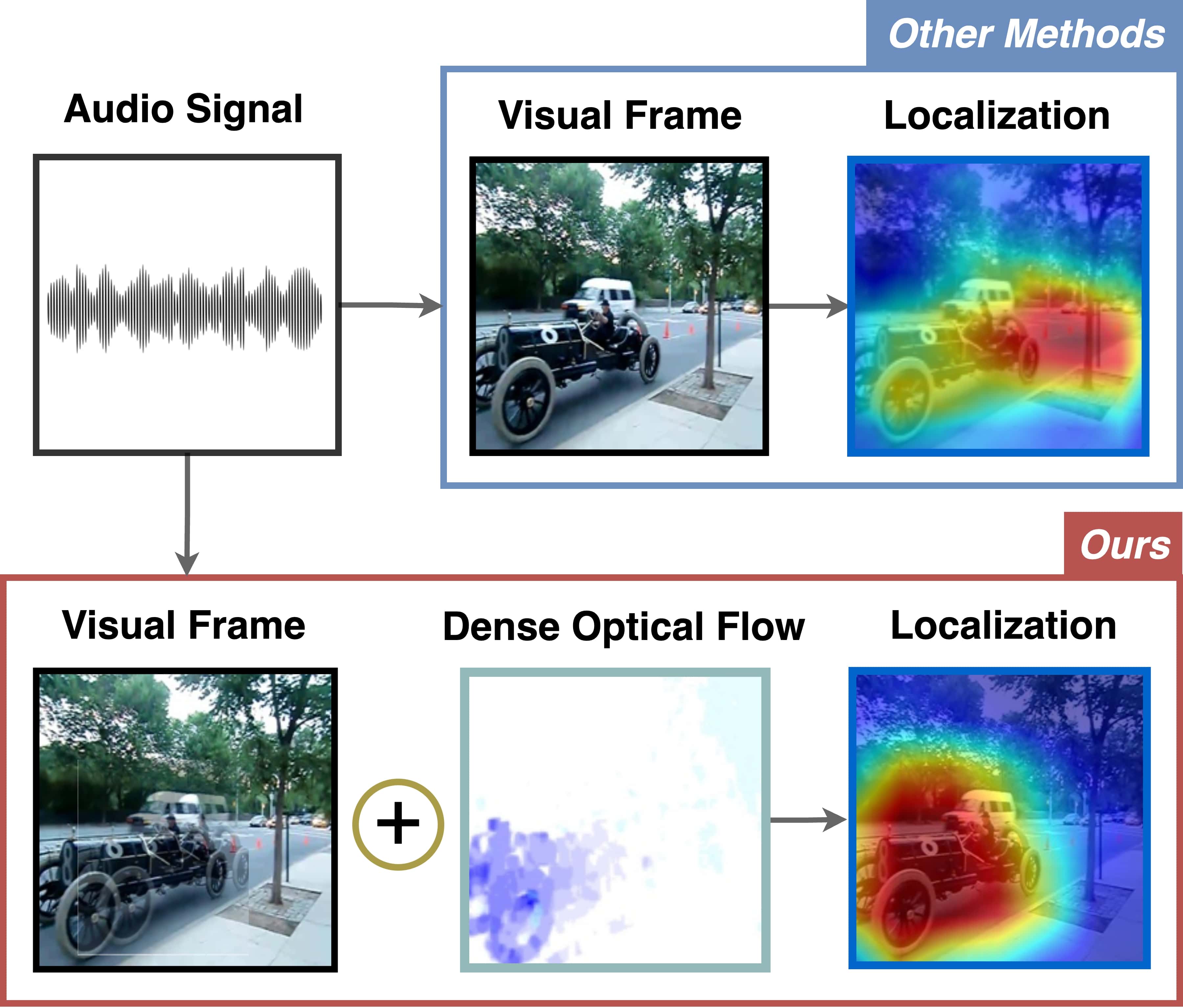}
\caption{Given a video with audio, the goal of sound source localization is to localize the object/region producing the sound in a video frame. Our method introduces \textit{optical flow} as an informative prior to improve visual sound source localization performance.}
\label{fig:conceptdiagram}
\vspace{-1.0em}
\end{figure}

Owing to the success of self-supervised learning (SSL) in vision \cite{chen2020simple, grill2020bootstrap}, language \cite{devlin2018bert, raffel2020exploring} and other multi-modal applications \cite{akbari2021vatt, mu2021slip}, recent methods in sound source localization \cite{chen2021localizing, song2022self} have adopted SSL based methods to overcome the need for annotations. One such method \cite{chen2021localizing}, finds the cosine similarity between the audio and visual representations extracted convolutionally at different spatial locations in the images. They rely on self-supervised training by creating positive and negative associations from these predicted similarity matrices. This bootstrapping approach has been shown to improve sound source localization. 

Following this research finding, the majority of recent approaches in visual sound source localization have focused on creating robust optimization objectives for better audio-visual associations. However, one interesting aspect of the problem that has received relatively little attention is the creation of \textit{informative priors} to improve the association of the audio to the correct ``sounding object" (or the object producing the sound). Priors can be viewed as potential regions in the image from where the sound may originate. We can draw parallels to work in two-stage object detection methods, in which region proposal networks are used to identify regions in the image space that could potentially be objects. However, generating potential candidate regions for sound source localization is more challenging because the generated priors should be relevant from a multi-modal perspective. In order to generate these informative priors for where sounds possibly originate from, we leverage optical flow. 

The intuition behind using optical flow to create an enhanced prior is the fact that optical flow can model patterns of apparent motion of objects. This is important as most often, an object moving in a video tends to be the sound source.  Enforcing a constraint to prioritize the objects that tend to be in relative motion might lend itself to creating better sound source localizations. This paper proposes an optical flow-based localization network that can create informative priors for performing superior sound source localization.  The contributions in this paper are as follows:

\begin{enumerate}
    \item We explore the need for creating informative priors for visual sound source localization, which is a complementary research direction to prior methods.
    \vspace{-2mm}
     \item We propose the use of optical flow as an additional source of information to create informative priors. 
    \vspace{-2mm}
    \item We design an optical flow-based localization network that uses cross-attention to form stronger audio-visual associations for visual sound source localization.
    \vspace{-2mm}
    \item We run extensive experiments on two benchmark datasets: VGG Sound and Flickr SoundNet and demonstrate the effectiveness of our method. Our method consistently achieves superior results over the state-of-the-art. We perform rigorous ablation studies and provide quantitative and qualitative results, showing the superiority of our novel localization network. 
\end{enumerate}

\section{Related Work}

Generating robust multi-modal representations through joint audio-visual learning is an active area of research that has found application in multiple audio-visual tasks. Initial works in the area of joint audio-visual learning focus on probabilistic approaches. In \cite{hershey1999audio}, the audio-visual signals were modeled as samples from a multivariate Gaussian process, and audio-visual synchrony was defined as the mutual information between the modalities. \cite{fisher2000learning} focused on first learning a lower-dimensional subspace that maximized mutual information between the two modalities. Furthermore, they explored the relationship between these audio-visual signals using non-parametric density estimators. \cite{izadinia2012multimodal} proposed a spatio-temporal segmentation mechanism that relies on the velocity and acceleration of moving objects as visual features and used canonical correlation analysis to associate the audio with relevant visual features. In recent years, deep learning-based methods have been used to explore the creation of better bimodal representations. They mostly employ two-stream networks to encode each modality individually and employ a contrastive loss-based supervision to align the two representations \cite{hu2019deep}. Methods like \cite{afouras2020self, zhao2019sound} used source separation to localize audio via motion trajectory-based fusion and synchronization. Furthermore, \cite {qian2020multiple}  addressed the problem of separating multiple sound sources from unconstrained videos by creating coarse to fine-grained alignment of audio-visual representations. Additionally, methods like \cite{owens2018audio, qian2020multiple} use class-specific saliency maps. \cite{zhou2016learning} uses class attention maps to help generate saliency maps that are used for better sound source localization. More recently, methods have focused on creating objective functions specific to sound localization. \cite{chen2021localizing} introduced the concept of tri-map, which incorporates background mining techniques into the self-supervised learning setting. The tri-map contains an area of positive correlation, no correlation (background), and an ignore zone to avoid uncertain areas in the visual space. \cite{song2022self} introduced a negative-free method for sound-localization by mining explicit positives. Further, this method uses a predictive coding technique to create better a feature alignment between the audio and visual modalities. These recent methods mainly focus on creating stronger optimization objectives for visual sound source localization. A complementary direction in the research landscape is to explore creating more informative priors for audio-visual association. In this paper, we explore one such idea, which leverages optical flow. The authors of \cite{arandjelovic2018objects} have explored the use of optical flow in the context of certain audio-visual tasks, like retrieval. In this work, we explore the use of optical flow as an informative prior for visual sound source localization. 

Optical flow provides a means to estimate pixel-wise motion between consecutive frames. Early works \cite{opticalflowearly2, opticalflowearly1,  opticalflowearly3} presented optical flow prediction as an energy minimization problem with several objective terms utilizing continuous optimization. Optical flow maps can be broadly divided into two types: Sparse and Dense. Sparse optical flow represents the motion of salient features in a frame, whereas Dense optical flow represents the motion flow vectors for the whole frame. Earlier methods for sparse optical flow estimation include the Lucas-Kanade algorithm \cite{lucaskanade}, which utilizes brightness constancy equations to optimize a least squares approximation under the assumption that flow remains locally smooth and the relative displacement of neighboring pixels is constant. Farneback \cite{farneback2003two} proposed a dense optical flow estimation technique where quadratic polynomials were utilized to approximate pixel neighborhood for two frames, and these polynomials were then used to compute global displacement. FlowNet \cite{flownet} proposed the first CNN-based approach towards estimating optical flow maps where they computed static cross-correlation between intermediate convolutional feature maps for two consecutive frames and up-scale them to extract optical flow maps.

\section{Method}

In this section, we will first present the formulation of the sound source localization problem under a supervised setting. Following this, we will describe the current self-supervised approach, motivate the need for better localization proposals for sound source localization, and subsequently elaborate on the design and implementation of our novel optical flow-based sound source localization network.

\subsection{Problem Statement} 

Given a video consisting of both the audio and visual modality, visual sound source localization aims to find the spatial region in the visual modality that generated the audio. Consider a video consisting of $N$ frames. Let the image corresponding to a video frame be $I$, where $ I \in \mathbb{R}^{W_i\text{x}H_i\text{x}3}$ and $A$ be the spectrogram representation generated out of the audio around the frame, where $A \in \mathbb{R}^{W_a\text{x}H_a\text{x}1}$.
The problem of audio localization can be thought of as finding the region in $I$ that has a high association/correlation with $A$. More formally, this can be written as: 
\begin{equation}
\label{feature_extactors}
\begin{split}
& f_v = \Phi(I;\theta_i)  ; f_a = \Psi(A;\theta_j) \\
& P(I,A) = \omega(f_v,f_a)
\end{split}
\end{equation}
where $\Phi(I;\theta_i)$ and $ \Psi(A;\theta_j)$ correspond to convolution neural network-based feature extractors associated with visual and audio modalities, and $f_v \in \mathbb{R}^{m\text{x}n\text{x}c} $ and $f_a \in \mathbb{R}^{m\text{x}n\text{x}c}$ are the corresponding lower dimensional feature maps, respectively. $\omega$ is the function that finds the association between the two modalities, and  $P(I, A)$ is the region in the original image space with the source that generated the audio. It is important to note that extrapolating the association in the feature space to the corresponding region of the original image space (i.e $P(I,A)$ ) is trivial. 
Given the above-mentioned feature maps, one way of finding an association between the feature representations is:
\begin{equation}
\label{supervised_Eq}
\begin{split}
& A_{avg} = GAP(f_a) \\
& S = \frac{f_v^i . A_{avg}}{||f_v^i||.||A_{avg}||},  \forall i \in [1,m*n] \\
\end{split}
\end{equation}

where $GAP(f_a)$ is the global-average-pooled representation of the audio feature map. S represents the cosine similarity of this audio representation to each spatial location in the visual feature map. Here $m$ and $n$ are the width and height of the feature map. If a binary mask $M \in \mathbb{R}^{m\text{x}n\text{x}1}$ generated from a ground truth indicating positive and negative regions of audio-visual correspondence is available, we can formulate the learning objective in a supervised setting:

For a given sample $k$ (with an image frame $I_k$ and audio $A_k$) in the dataset, the positive and negative response can be defined as 
\begin{equation}
\begin{split}
& \text{Pos}_k = \frac{1}{|{M_k}|} \langle M_k, S_ {k \rightarrow k} \rangle \\
& \text{Neg}_k = \frac{1}{|1-{M_k}|} \langle 1-M_k , S_ {k \rightarrow k} \rangle + \frac{1}{m*n}\underset{k\neq j}{\sum} \langle 1, S_{k \rightarrow j} \rangle \\ 
\end{split}
\end{equation}
Here $S_ {k \rightarrow k}$ refers to the cosine similarity $S$ from Eq \ref{supervised_Eq} when using $I_k$ and $A_k$. Similarly, $S_{k \rightarrow j}$ is the cosine similarity when the image and audio are not from the same video. $\langle\cdot,\cdot\rangle$ denotes the inner product. The final learning objective has a similar formulation to \cite{oord2018representation}:
\begin{equation}
\label{objective}
\begin{split}
& L = -\underset{\pazocal{X}}{\sum} \Bigg\{ \text{log}\bigg ( \frac{\exp{(Pos_k)}}{\exp{(Pos_k)} +\exp{(Neg_k)}} \bigg)\bigg\}
\end{split}
\end{equation}

\begin{figure*}[t!]
\centering
\includegraphics[scale=0.2]{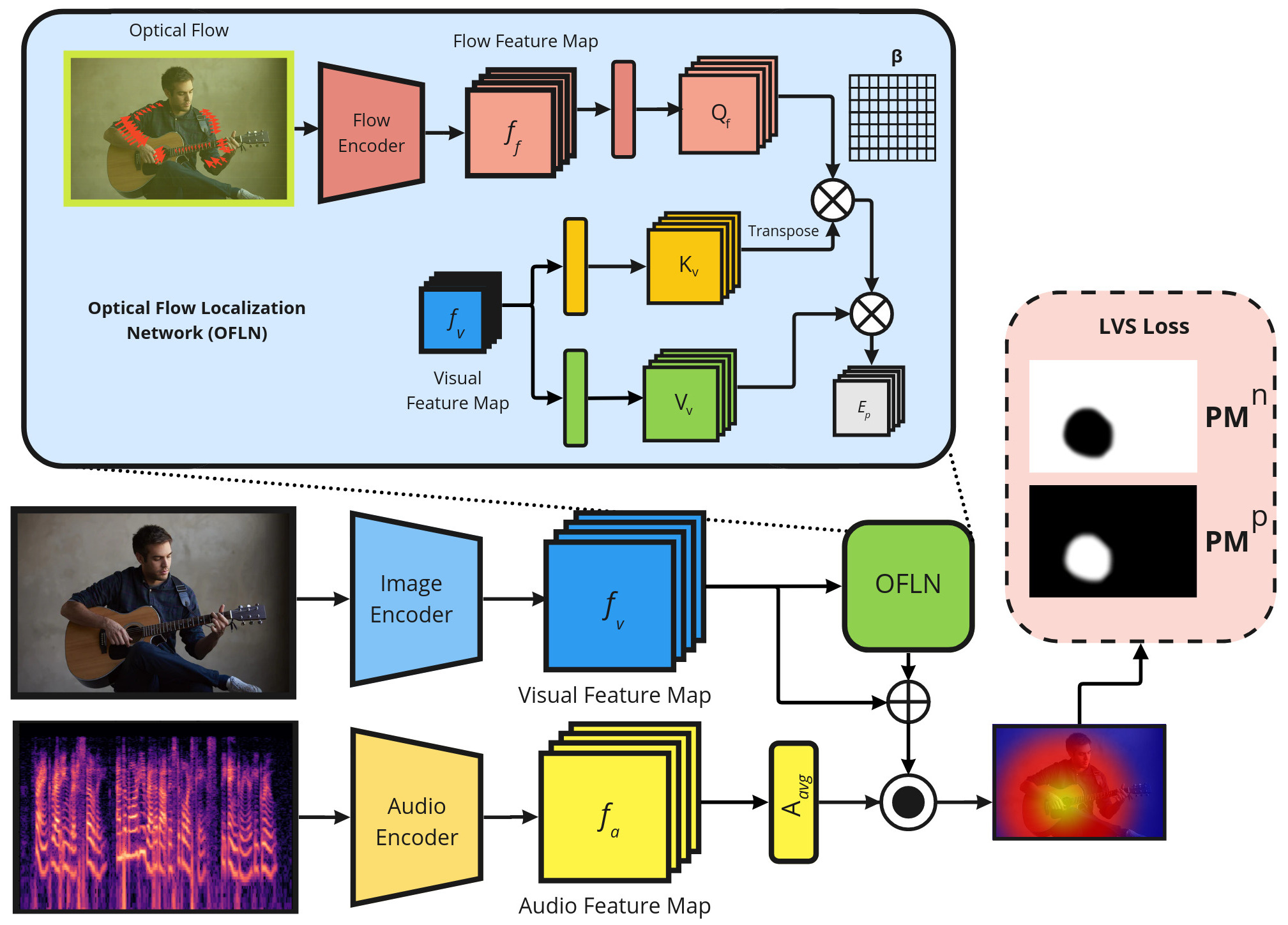}
\caption{Overview of our optical flow-based sound localization method. Given a chosen frame of a video and the audio surrounding that frame, we extract features from both modalities, which are then used to attend towards sounding objects in the frame. We further compute the dense optical flow field from the chosen and subsequent frame and use flow features to attend towards moving objects in the frame.}
\label{fig:deeppipeline}
\vspace{-0.5em}
\end{figure*}

\subsection{Self-Supervised Localization}
\label{sec:ssl}

In most real-world scenarios, the ground truth necessary to generate the binary mask $M$ would be missing. Hence there is a need for a training objective that does not rely on explicit ground truth annotations. One way to achieve this objective is to replace the ground truth mask with a generated pseudo mask as proposed in \cite{chen2021localizing}. The pseudo mask can be generated by binarizing the similarity matrix $S$ based on a threshold. More specifically, given $S_ {k \rightarrow k}$ from Eq \ref{supervised_Eq}, the pseudo mask can be written as:
\begin{equation}
\label{pseudomask}
\begin{split}
& PM  = \sigma(S_ {k \rightarrow k}-\epsilon)/\tau \\
\end{split}
\end{equation}
where $\epsilon$ is a scalar threshold. $\sigma$ denotes the sigmoid function that maps a similarity value in $S_{k \rightarrow k}$, that is below the threshold to value 0 and above the threshold to 1. $\tau$ is the temperature controlling the sharpness. Additionally, \cite{chen2021localizing} further refines the pseudo mask by eliminating potentially noisy associations. This is done by considering separate positive and negative thresholds above and below the similarity value that is considered reliable. If a value is between these thresholds, it's considered a noisy association and is subsequently ignored. More formally:
\begin{equation}
\label{self_supervised}
\begin{split}
& \text{PM}_k^p  = \sigma(S_ {k \rightarrow k}-\epsilon_p)/\tau \\
& \text{PM}_k^n  = \sigma(S_ {k \rightarrow k}-\epsilon_n)/\tau \\
& \text{Pos}_k = \frac{1}{|{PM_k^p}|} \langle PM_k^p, S_ {k \rightarrow k} \rangle \\
& \begin{aligned} \text{Neg}_k = \frac{1}{|1-{PM_k^n}|} \langle 1-PM_k^n , S_ {k \rightarrow k} \rangle\\   +  \frac{1}{m*n}\underset{k\neq j}{\sum} \langle 1,
S_{k \rightarrow j} \rangle \end{aligned} \\ 
\end{split}
\end{equation}
Here $\epsilon_p$ and $\epsilon_n$ are the positive and negative thresholds, respectively. Once the positive and negative responses are computed, the overall training objective is similar to Eq \ref{objective}.

In the above approach, it is logical to bootstrap the prediction and perform self-supervised training if the pseudo masks in Eq \ref{pseudomask} generated at the initial training iterations resemble that of the ground truth. However, this is not guaranteed since the feature extractors associated with the individual modalities (in Eq  \ref{feature_extactors}) are randomly initialized. Therefore, a high or low value in the similarity matrix $S_{k \rightarrow k}$ during the initial iterations of the self-supervised training may not correspond to informative positive or negative regions since the feature extractors are not trained. If a feature extractor is initialized with pretrained weights from a classification task, for example, the visual extractor on ImageNet, the network will often activate towards objects in the image. Considering this characteristic as an \textit{object-centric} prior, it may be useful for self-supervised sound localization as the most salient objects in a frame are often the ones emitting the sound. However, situations may arise where the source of the audio is not the most salient object in the frame. This would produce sub-optimal associations $S_ {k \rightarrow k}$ in the initial iterations, which, when used for self-supervised training as mentioned in Eq \ref{self_supervised} would lead to sub-optimal performance. As a result, there is a need to construct more meaningful priors when computing $S_{k \rightarrow k}$ to improve audio-visual associations, subsequently improving self-supervised learning.

\subsection{Optical-Flow Based Localization Network}
Having motivated the need for some meaningful priors that enable better audio-visual associations, we approach the problem from an object detection viewpoint. In earlier object detection methods such as R-CNN \cite{girshick2014rich} and Fast R-CNN \cite{girshick2015fast}, a selective search was used as a method to generate region proposals. Selective search provided a set of probable locations where an object of interest may be present. An alternative to selective search-based approaches is a two-stage approach like in \cite{ren2015faster}, using region proposal networks. Most of these region proposal networks have auxiliary training objectives in order to produce regions containing potential objects. Using these objectives to generate potential regions of interest in a self-supervised setting becomes challenging. Furthermore, generating candidate regions using selective search or regular region proposal networks, only based on visual modality, might not be well suited for enforcing priors for a cross-modal task such as visual sound source localization. 

As a better alternative, we use optical flow to generate informative localization proposals. Optical flow using frames of a video can efficiently capture the objects that are moving. Most often, these objects are the source of the sound. Capturing optical flow in the pixel space can often be a good prior to improve audio-visual association. Furthermore, since the optical flow tends to focus on the relative motion of objects rather than the salient objects, it can complement the priors of the pre-trained vision model, which tends to focus on the latter. We design a network as shown in Figure \ref{fig:deeppipeline}, which takes in optical flow computed between two adjacent video frames and generates regions in the feature map $f_v$ that act as priors to create better audio-visual associations. The localization network is comprised of a cross-attention between the feature representation extracted from image and flow modalities. Given the flow feature representation $f_f$ and visual feature representation $f_v$, we project these feature representations using separate projection layers to create two tensors $K_v$ and $Q_f$. $\beta$ is computed as an outer product of the tensors $K_v$ and $Q_f$ along the channel dimensions. That is, if $ K_v$ and $Q_f$ $ \in \mathbb{R}^{m\text{x}n\text{x}d} $, then the resulting $\beta$ $ \in \mathbb{R}^{m\text{x}n\text{x}d\text{x}d} $ is computed as below:
\begin{equation}
\label{attention}
\begin{split}
\beta = softmax(\frac{K_v \odot Q_f}{\sqrt{d}} )
\end{split}
\end{equation}
The softmax function is applied to the final dimension to normalize the attention matrix. The goal is to compute the attention to be applied to each spacial location, thus yielding a cross attention matrix of size $d \text{x} d$ for each spatial location. We compute another tensor from the visual modality ${V_v}$ $ \in \mathbb{R}^{m\text{x}n\text{x}d}$.  For each spatial location in $V_v$, we have a $d$ dimensional representation which we multiply to the corresponding $d \text{x} d$ attention matrix in $\beta$. That is :
\begin{equation}
\label{attention2}
E = V_v^{ij}\beta^{ij};   \forall i \in [1,m]; \forall j \in [1,n]
\end{equation}

Finally $E$ is projected back to produce the final cross-attended proposal prior $E_p$ $\in \mathbb{R}^{m\text{x}n\text{x}c}$.  In order to impose this prior for performing the audio-visual association, we add $E_p$ to the visual feature map $f_v$ as shown in Figure \ref{fig:deeppipeline}. The enhanced audio-visual association can be written as:
\begin{equation}
\begin{split}   
& f_{enh} = f_v \oplus E_p \\
& S^{enh}  = \frac{f_{enh}^i . A_{avg}}{||f_{enh}^i||.||A_{avg}||},  \forall i \in m*n \\
\end{split}
\end{equation}
where $\oplus$ denotes element-wise addition. Once the enhanced audio-visual association is obtained, we use Eq \ref{self_supervised} to compute the positive and negative responses. We train the entire network (feature extractors and localization network) end-to-end using the optimization objective mentioned in  Eq \ref{objective}.

\section{Experiments}

\subsection{Datasets}
For training and evaluating our proposed model, we follow prior work in this area and use two large-scale audio-visual datasets:

\subsubsection{Flickr SoundNet} 

Flickr SoundNet \cite{aytar2016soundnet} is a collection of over 2 million unconstrained videos collected from the Flickr platform. To directly compare against prior works, we construct two subsets of 10k and 144k videos that are preprocessed into extracted image-audio pairs, described further in Section \ref{sec:impdetails}. The Flickr SoundNet evaluation dataset consists of 250 image-audio pairs with labeled bounding boxes localized on the sound source in the image, manually annotated by \cite{senocak2018learning}. 

\subsubsection{VGG Sound} 

VGG Sound \cite{chen2020vggsound} is a dataset of 200k video clips spread across 309 sound categories. Similar to Flickr SoundNet, we construct subsets of 10k and 144k image-audio pairs to train our proposed model. For evaluation, we utilize the VGG Sound Source \cite{chen2021localizing} dataset, which contains 5000 annotated image-audio pairs that span 220 sound categories. Compared to the Flickr SoundNet test set, which has about 50 sound categories, VGG Sound Source has significantly more sounding categories, making it a more challenging scenario for sound localization.

\subsection{Evaluation Metrics}

For proper comparisons against prior works, we use two metrics to quantify audio localization performance: Consensus Intersection Over Union (cIoU) and Area Under Curve of cIoU scores (AUC) \cite{senocak2018learning}. cIoU quantifies localization performance by measuring the intersection over union of a ground-truth annotation and a localization map, where the ground-truth is an aggregation of multiple annotations, providing a single consensus. AUC is calculated by the area under the curve of cIoU created from thresholds varying from 0 to 1. In our experiments, we show results for cIoU at a threshold of 0.5, denoted by $cIoU_{0.5}$, and AUC scores, denoted by $AUC_{cIoU}$.

\subsection{Implementation Details}
\label{sec:impdetails}

In this paper, sound source localization is defined as localizing an excerpt of audio to its origin location in an image frame, both extracted from its respective video clip. For both Flickr SoundNet and VGG Sound, we extract the middle frame of a video along with 3 seconds of audio centered around the middle frame and a calculated dense optical flow field to construct an image-flow-audio pair. For the image frames, we resize images to $224 \times 224$ and perform random cropping and horizontal flipping data augmentations. To calculate an optical flow field corresponding to the middle frame, we take the middle frame and subsequent frame of a video $V$, denoted by $V_{t}$ and $V_{t+1}$ respectively, and use the Gunnar Farneback \cite{farneback2003two} algorithm to generate a 2-channel flow field corresponding to horizontal and vertical flow vectors denoting movement magnitude. We similarly perform random cropping and horizontal flipping of the flow fields, which are performed consistently with image augmentations. For audio, we sample 3 seconds of the video at 16kHz and construct a log-scaled spectrogram using a bin size of 256, FFT window of 512 samples, and stride of 274 samples, resulting in a shape of $257 \times 300$.

Following \cite{chen2021localizing}, we use ResNet18 backbones as the visual and audio feature extractors. Similarly, we use ResNet18 as the optical flow feature extractor. We pretrain the visual and flow feature extractors on ImageNet and leave the audio network randomly initialized. During training, we keep the visual feature extractor parameters frozen. For all experiments, we train the model using the Adam optimizer with a learning rate of $10^{-3}$ and a batch size of 128. We train the model for 100 epochs for both the 10k and 144k sample subsets on Flickr SoundNet and VGGSound. We set $\epsilon_{p}=0.65$, $\epsilon_{n}=0.4$, and $\tau=0.03$, as described in Eq \ref{self_supervised}.

\begin{table}[]
\setlength{\tabcolsep}{0.3em}
\begin{center}
\begin{tabular}{lccc}
\hline
Method       & Training Set                 & \multicolumn{1}{l}{cIoU$_{0.5}$} & \multicolumn{1}{l}{AUC$_{cIoU}$} \\ \hline
Attention \cite{senocak2018learning}   & \multirow{6}{*}{Flickr 10k}  & 0.436                             & 0.449                             \\
CoarseToFine \cite{qian2020multiple} &                              & 0.522                             & 0.496                             \\
AVObject \cite{afouras2020self}    &                              & 0.546                             & 0.504                             \\
LVS$^*$ \cite{chen2021localizing}         &                              & 0.730                             & 0.578                             \\
SSPL \cite{song2022self}        &                              & 0.743                             & 0.587                             \\
\textbf{HTF (Ours)}   &                              & \textbf{0.860}                    & \textbf{0.634}                    \\ \hline
Attention \cite{senocak2018learning}    & \multirow{7}{*}{Flickr 144k} & 0.660                             & 0.558                             \\
DMC \cite{hu2019deep}         &                              & 0.671                             & 0.568                             \\
LVS$^*$ \cite{chen2021localizing}         &                              & 0.702                             & 0.588                        \\
LVS$^\dagger$ \cite{chen2021localizing} & & 0.697 & 0.560 \\
HardPos \cite{senocak2022learning}     &                              & 0.762                             & 0.597                             \\
SSPL \cite{song2022self}         &                              & 0.759                             & 0.610                             \\
\textbf{HTF (Ours)}   &                              & \textbf{0.865}                    & \textbf{0.639}                    \\ \hline

LVS$^*$ \cite{chen2021localizing} & \multirow{4}{*}{VGGSound 144k} & 0.719 & 0.587 \\
HardPos \cite{senocak2022learning} & & 0.768 & 0.592 \\
SSPL \cite{song2022self} & & 0.767 & 0.605 \\
\textbf{HTF (Ours)} & & \textbf{0.848} & \textbf{0.640} \\ \hline

\end{tabular}
\end{center}
\caption{Quantitative results on the Flickr SoundNet testing dataset where models are trained on the two training subsets of Flickr SoundNet and VGG Sound 144k. ``*" Denotes our faithful reproduction of the method, and ``$\dagger$" denotes our evaluation reproduction using officially provided model weights.}
\label{tab:flickrresult}
\end{table}

\begin{table}[]
\setlength{\tabcolsep}{0.4em}
\begin{center}
\begin{tabular}{lccc}
\hline
Method     & Training Set                   & \multicolumn{1}{l}{cIoU$_{0.5}$} & \multicolumn{1}{l}{AUC$_{cIoU}$} \\ \hline
Attention \cite{senocak2018learning}  & \multirow{4}{*}{VGGSound 10k}  & 0.160                             & 0.283                             \\
LVS$^*$ \cite{chen2021localizing}       &                                & 0.297                             & 0.358                             \\
SSPL \cite{song2022self}       &                                & 0.314                             & 0.369                             \\
\textbf{HTF (Ours)} &                                & \textbf{0.393}                    & \textbf{0.398}                    \\ \hline
Attention \cite{senocak2018learning}  & \multirow{7}{*}{VGGSound 144k} & 0.185                             & 0.302                             \\
AVObject \cite{afouras2020self}   &                                & 0.297                             & 0.357                             \\
LVS$^*$ \cite{chen2021localizing}        &                                & 0.301                             & 0.361                             \\
LVS$^\dagger$ \cite{chen2021localizing} & & 0.288 & 0.359 \\
HardPos \cite{senocak2022learning}   &                                & 0.346                             & 0.380                             \\
SSPL \cite{song2022self}      &                                & 0.339                             & 0.380                             \\
\textbf{HTF (Ours)} &                                & \textbf{0.394}                    & \textbf{0.400}                    \\ \hline
\end{tabular}
\end{center}
\caption{Quantitative results on the VGG Sound Source testing dataset where models are trained on the two training subsets of VGG Sound.}
\label{tab:vggresult}
\end{table}

\begin{table}[]
\setlength{\tabcolsep}{0.3em}
\begin{center}
\begin{tabular}{lccc}
\hline
Method              & Testing Set                        & \multicolumn{1}{l}{cIoU$_{0.5}$} & \multicolumn{1}{l}{AUC$_{cIoU}$} \\ \hline
LVS$^*$ \cite{chen2021localizing}                  & \multirow{2}{*}{VGGSS Heard 110}   & 0.251                             & 0.336                             \\
\textbf{HTF (Ours)} &                                    & \textbf{0.373}                    & \textbf{0.386}                    \\ \hline
LVS$^*$ \cite{chen2021localizing}                  & \multirow{2}{*}{VGGSS Unheard 110} & 0.270                             & 0.349                             \\
\textbf{HTF (Ours)} &                                    & \textbf{0.393}                    & \textbf{0.400}                    \\ \hline
\end{tabular}
\end{center}
\caption{Quantitative results on the VGG Sound Source testing dataset on heard and unheard class subsets. Each model is trained on 50k samples belonging to 110 (heard) classes.}
\label{tab:heardunheard}
\end{table}

\begin{figure*}[t!]
\centering
\includegraphics[scale=0.087]{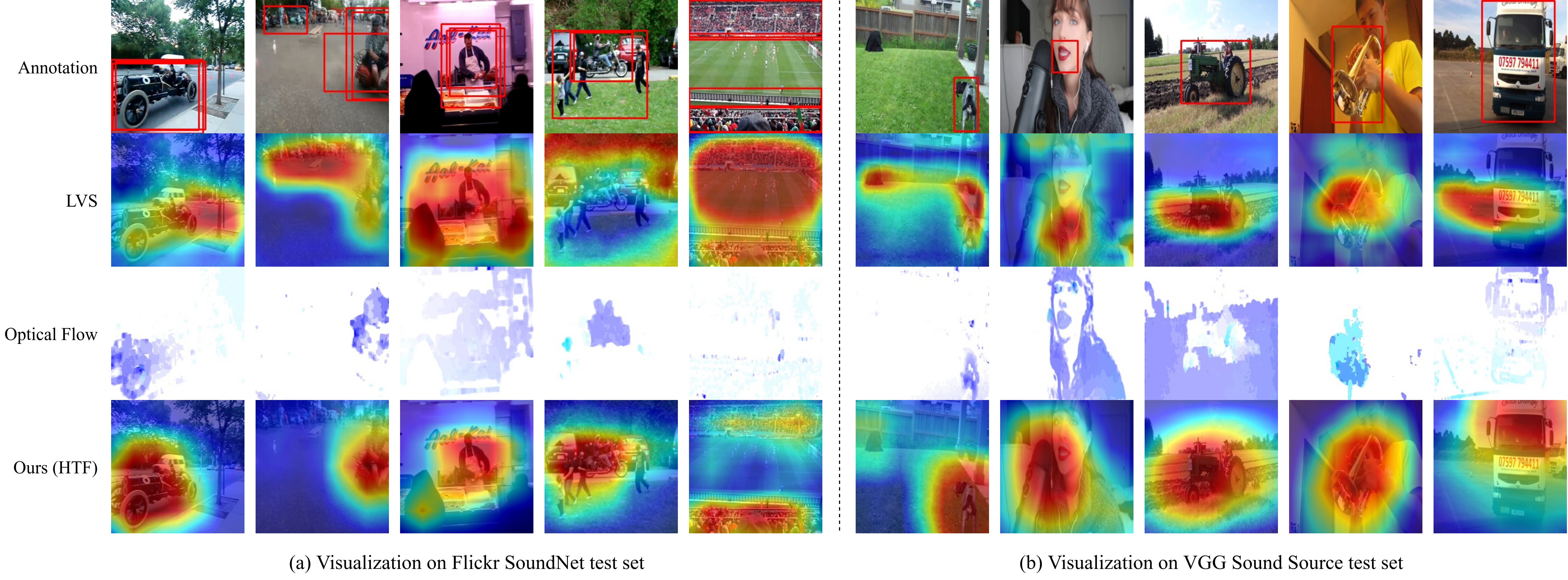}
\caption{Qualitative results of our method on each testing set. Example visualizations are from models trained on each test set's respective 144k training set. Our method effectively localizes towards sounding objects exhibiting movement. Figure is best viewed in color. }
\label{fig:qual}
\end{figure*}

\begin{figure}
\includegraphics[scale=0.09]{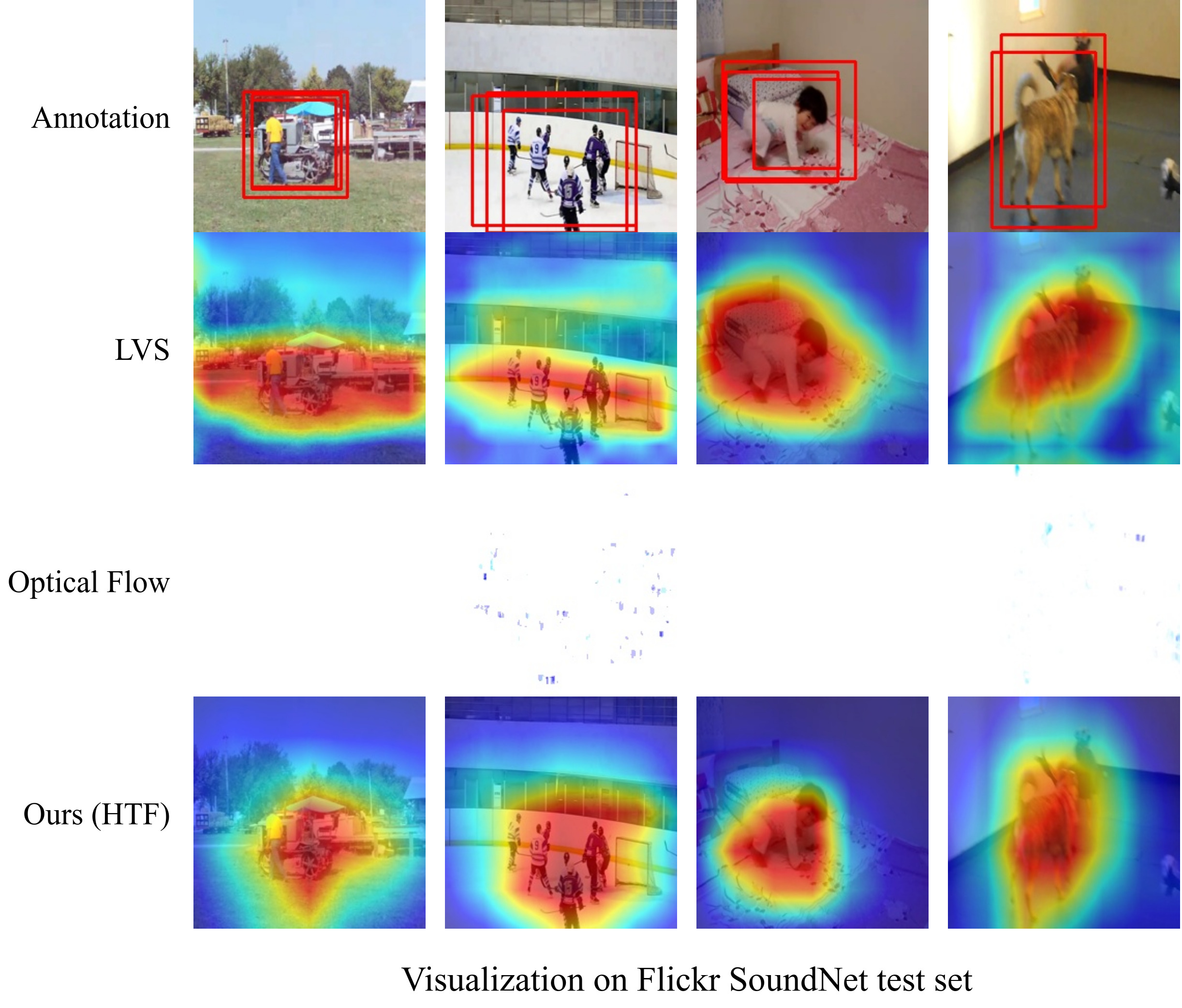}
\caption{Qualitative results of our method on the Flickr SoundNet test set. Examples are from models trained on the Flickr 144k set. Even in the absence of meaningful optical flow, our method can still localize to the sound source. Figure is best viewed in color.}
\label{fig:qual2}
\end{figure}

\subsection{Quantitative Evaluation}

In this section, we compare our method against prior works \cite{afouras2020self, chen2021localizing, hu2019deep, qian2020multiple, senocak2018learning, senocak2022learning, song2022self} on standardized experiments for self-supervised visual sound source localization. Results of various training configurations are reported in Tables \ref{tab:flickrresult} and \ref{tab:vggresult} for the Flickr SoundNet and VGG Sound Source testing datasets, respectively. 

As shown in Table \ref{tab:flickrresult} and \ref{tab:vggresult}, our method, HTF, significantly outperforms all prior methods, creating the new state-of-the-art on self-supervised sound source localization. On the Flickr testing set, we achieved an improved performance of 11.7\% cIoU and 4.7\% AUC when trained on 10k Flickr samples and 10.6\% cIoU and 2.9\% AUC when trained with 144k Flickr samples. Similarly, on the VGG Sound Source testing set, we improve by 7.9\% cIoU and 2.9\% AUC when trained on 10k VGG Sound samples and 5.5\% cIoU and 2.0\% AUC when trained on 144k samples.

Further, we investigate the robustness of our method by evaluating it across the VGG Sound and Flickr SoundNet datasets. Specifically, we train our model with 144k VGG Sound samples and test on the Flickr SoundNet test set. Comparing against \cite{chen2021localizing, senocak2022learning, song2022self}, we significantly outperform all methods, as shown in Table \ref{tab:flickrresult}, which shows our model is capable of generalizing well across datasets. We further investigate our method's robustness by testing on sound categories that are \textit{disjoint} from what is seen during training. Following \cite{chen2021localizing}, we sample 110 sound categories from VGG Sound for training and test on the same 110 categories (heard) used during training and 110 other disjoint (unheard) sound categories. As shown in Table \ref{tab:heardunheard}, we outperform \cite{chen2021localizing} on both heard and unheard testing subsets. In addition, we highlight that the performance of the unheard subset slightly outperforms the heard subset, showing our model performs well against unheard sound categories.  

Utilizing a self-supervised loss formulation similar to \cite{chen2021localizing}, we see that our method significantly outperforms it on both testing datasets across all training setups and experiments. We highlight that these improvements are obtained from incorporating a more informative prior, based on optical flow, into the sound localization objective. In section \ref{sec:ablstudy}, we further investigate the direct influence of incorporating optical flow along with our other design choices.

\subsection{Qualitative Evaluation}

In Figure \ref{fig:qual}, we visualize and compare sound localizations of LVS \cite{chen2021localizing} and our method on the Flickr SoundNet and VGG Sound Source test sets. As shown, our method can accurately localize various types of sound sources. Comparing against LVS \cite{chen2021localizing}, we examine localization improvements across multiple samples, specifically where sounding objects exhibit a high flow magnitude through movement. For example, in the first column, LVS \cite{chen2021localizing} localizes only a small portion of the sounding vehicle, while our method entirely localizes the vehicle, where a significant magnitude of flow is exhibited. In the fifth column, our method more accurately localizes to the two crowds in the stadium, both of which are sound sources exhibiting movement.

However, it is also important to investigate samples where little optical flow is present. It is possible that a frame in a video exhibits little movement, for example, a stationary car or person emitting noise. In these cases, there is no meaningful optical flow to localize towards. In Figure \ref{fig:qual2}, we see that even in the absence of significant optical flow, our method still localizes on par or better compared to LVS \cite{chen2021localizing}. This reinforces that optical flow is used as an \textit{optional} prior, where areas of high movement when present, \textit{may} be used to localize better, but is not required. In the following section, we further investigate the exact effects of introducing priors like optical flow into the self-supervised framework.

\subsection{Ablation Studies}
\label{sec:ablstudy}

In this section, we explore the implications of our design choices with multiple ablation studies. As explained in section \ref{sec:ssl}, we explore the need for informative priors to train a self-supervised audio localization network. We introduce optical flow as one of these priors, in addition to pretraining the vision network on ImageNet to provide an object-centric prior. In Table \ref{tab:ablstudy1}, we study the individual effects of each of these design choices, namely adding the flow attention mechanism, ImageNet weights for the vision encoder, and freezing the vision encoder during training. 

When training the model without any priors (model 4.a), we see that performance suffers as there is little meaningful information for the self-supervised objective. However, when simply adding the optical flow attention previously described (model 4.d), we see a large performance improvement as the network can now use optical flow to better localize, as moving objects are often the ones exhibiting sound. Similarly, when using ImageNet pretrained weights (Table 4.b), we see a significant performance improvement as the model now has an object-centric prior, where salient objects in an image are often exhibiting sound. When combining both priors (model 4.e), we see even further performance improvements, which show the importance of incorporating \textit{multiple} informative priors for the self-supervised sound localization objective.

\begin{table}[]
\setlength{\tabcolsep}{0.4em}
\begin{center}
\begin{tabular}{cccccc}
\hline
Model & Flow                  & \begin{tabular}[c]{@{}c@{}}Pretrain\\ Vision\end{tabular} & \begin{tabular}[c]{@{}c@{}}Frozen\\ Vision\end{tabular} & cIoU$_{0.5}$   & AUC$_{cIoU}$   \\ \hline
a & $\times$              & $\times$                                                  & $\times$                                                & 0.129          & 0.275          \\
b & $\times$              & $\checkmark$                                              & $\times$                                                & 0.315          & 0.364          \\
c & $\times$              & $\checkmark$                                              & $\checkmark$                                            & 0.306          & 0.362          \\ \hline
d & $\checkmark$          & $\times$                                                  & $\times$                                                & 0.271          & 0.343          \\
e & $\checkmark$          & $\checkmark$                                              & $\times$                                                & 0.382         & 0.394          \\
\textbf{f} & \textbf{$\checkmark$} & \textbf{$\checkmark$}                                     & \textbf{$\checkmark$}                                   & \textbf{0.393} & \textbf{0.398} \\ \hline
\end{tabular}
\end{center}
\caption{Ablation study on incorporating optical flow and strategies on the vision encoder. All models are trained on VGG Sound 10k and tested on the VGG Sound Source test set.}
\label{tab:ablstudy1}
\end{table}

We further explore the effects of freezing the vision encoder during training. As previously mentioned, a network pretrained on a classification task such as ImageNet will often have high activations around salient objects (an object-centric prior). When training in a self-supervised setting, the network may divert from its original weights and instead have a less object-centric focus, which may be suboptimal for sound source localization. When freezing the network in the non-flow setting (model 4.c), we see performance slightly decrease compared to the unfrozen counterpart (model 4.b). However, when freezing the network in the optical flow setting (model 4.f), we see a slight improvement over the flow setting with an unfrozen vision encoder (model 4.e). We infer that enforcing the vision encoder to keep its object-centric characteristics while the flow encoder can reason and attend towards other parts of the image produces a more informative representation, improving localization performance. 

Finally, we explore variations of the optical flow encoder to better understand how the optical flow information is being used. We replace the learnable ResNet18 encoder with a single max pooling layer to see if the simple presence of movement is still informative for localizing sounds. As shown in Table \ref{tab:ablstudy2}, when using a simple max pooling layer (model 5.a), we still notice a significant performance improvement over the network without optical flow (models 4.a-c). However, we see a further improvement over the max pooling layer when using a learnable encoder, like a ResNet18 network. While the max pooling layer only captures the presence of movement at a particular location, a learnable encoder allows deeper reasoning of the flow information. For example, the eighth column in Figure \ref{fig:qual} shows an optical flow field where the sounding object (tractor) is not moving but rather the environment around it is. In this case, with the max pooling encoder, the network is biased away from the sounding object, whereas a learnable encoder can better reason about the flow in the given frame, improving overall localization performance.

\begin{table}[]
\setlength{\tabcolsep}{0.3em}
\begin{center}
\begin{tabular}{ccccc}
\hline
Model & Flow Network      & Training Set                    & cIoU$_{0.5}$   & AUC$_{cIoU}$   \\ \hline
a & MaxPool           & \multirow{2}{*}{VGGS 10k}  & 0.379          & 0.393          \\
\textbf{b} & \textbf{ResNet18} &                                 & \textbf{0.393} & \textbf{0.398} \\ \hline
c & MaxPool           & \multirow{2}{*}{VGGS 144k} & 0.381          & 0.393          \\
\textbf{d} & \textbf{ResNet18} &                                 & \textbf{0.394} & \textbf{0.400} \\ \hline
\end{tabular}
\end{center}
\caption{Ablation study on configurations of the flow encoder network across training settings. All models are tested on the VGG Sound Source test set.}
\label{tab:ablstudy2}
\end{table}

\section{Conclusion}

In this work, we introduce a novel self-supervised sound source localization method that uses optical flow to aid in the localization of a sounding object in a frame of a video. In a video, moving objects are often the ones making the sound. We take advantage of  this observation by using optical flow as a prior for the self-supervised learning setting. We formulate the self-supervised objective and describe the cross-attention mechanism of optical flow over the corresponding video frame. We evaluate our approach on standardized datasets and compare against prior works and show state-of-the-art results across all experiments and evaluations. Further, we conduct extensive ablation studies to show the necessity and effect of including informative priors like optical flow, into the self-supervised sound localization objective to improve performance. 

While we explore optical flow in this work, there are other priors that may be explored to improve sound source localization further. For example, pretraining the audio encoder can likely provide a better understanding of the \textit{class} of the sound being emitted, which can then be used to help localize toward that specific object. Further, improving the optical flow generation, for example, using flow estimation methods or aggregating flow across multiple frames, can potentially improve the optical flow signal to ultimately improve overall localization performance. We leave exploration of these hypotheses for future work. \vspace{8px} \\ 
\textbf{Acknowledgements:} This work was supported by the Center for Identification Technology Research (CITeR) and the National Science Foundation (NSF) under grant \#1822190.

\newpage

{\small
\bibliographystyle{ieee_fullname}
\bibliography{egbib}
}

\newpage
~\newpage

\appendix

\section{Training And Evaluation Procedure}

\subsection{Dataset Construction}
\label{dataconstruction}

As mentioned in the main paper, we train our method on the VGG Sound \cite{chen2020vggsound} and Flickr SoundNet \cite{aytar2016soundnet} datasets. Both of these datasets are originally aggregated from online video streaming services, YouTube and Flickr, respectively. The authors of both datasets release only video IDs hosted on the platform, meaning that users of these datasets need to individually download and preprocess videos. A portion of these videos have become unrecoverable, as they have either been removed or blocked on the platform.

To construct the training sets of both datasets, we download a random subset of videos in each dataset until enough available videos are collected to construct the 10k and 144k training sets. We then preprocess the videos, extract audio, and construct optical flow maps as described in the main paper. 

For the VGG Sound Source \cite{chen2021localizing} testing set, 5,158 YouTube IDs are provided as the official testing set. At the time of dataset construction, 488 videos are unrecoverable, resulting in a testing set of 4,670 samples. We use these available samples as the VGG Sound Source testing set to construct image-flow-audio pairs for evaluating our method. For the Flickr SoundNet test set, \cite{senocak2018learning} provides 250 preprocessed image-audio testing pairs that are directly available from the author's official project page\footnote{\url{https://github.com/ardasnck/learning_to_localize_sound_source}}. However, \cite{senocak2018learning} does not provide the original videos of these 250 testing samples, which are required for constructing optical flow fields. We recover these original videos from the Flickr platform, find the corresponding video frame that the original test sample of that video belongs to, and construct an optical flow field using the subsequent frame. In this process, we are able to recover 178 videos, making our Flickr SoundNet consist of 178 image-flow-audio samples.  

\subsection{Expanded Flickr SoundNet Test Set}
\label{sec:expand}

The Flickr SoundNet test set, created by \cite{senocak2018learning}, contains 250 annotated samples randomly selected out of 2,786 total annotated samples. \cite{senocak2018learning} originally used these annotated samples to explore supervised and semi-supervised learning methods for visual sound source localization. With the current research landscape of self-supervised sound source localization, the 250 testing samples are used for evaluation and the remaining annotated samples provided by \cite{senocak2018learning} are disregarded. Since our method optimizes a self-supervised objective with no explicit annotations for training, one such alternative is to use these remaining annotated samples for expanding the Flickr SoundNet test set, as they are otherwise unused. 

We collect and preprocess these remaining annotated samples to construct a novel Flickr SoundNet test set consisting of 1,769 samples. As shown in Table \ref{tab:expandedflickrresult}, we show results on the expanded Flickr SoundNet test set in a similar fashion to the quantitative results in the main paper on the official Flickr SoundNet test set. Specifically, we compare our method against LVS \cite{chen2021localizing}, where models are trained on both subsets of the Flickr SoundNet training sets, in addition to VGG Sound 144k.

As shown, our method significantly outperforms LVS \cite{chen2021localizing} in the expanded testing scenario, showing our method is still robust to a much larger-scale testing set that spans more sounding categories than the official test set. In addition, when comparing against testing on the official 250 samples, we see a reduction in performance across all methods, showing that expanding the testing set leads to a more challenging sound source localization scenario. For example, for our method trained on Flickr 144k, we achieve 0.865 cIoU and 0.639 AUC on the official test set, compared to 0.759 cIoU and 0.575 AUC on the expanded testing set we introduce. We believe this evaluation on 1,769 annotated samples instead of 250 samples offers a more robust and representative testing set, which can be used for future self-supervised visual sound localization works for improved evaluations. 

\begin{table}[]
\setlength{\tabcolsep}{0.5em}
\begin{center}
\begin{tabular}{lccc}
\hline
Method       & Training Set                 & \multicolumn{1}{l}{cIoU$_{0.5}$} & \multicolumn{1}{l}{AUC$_{cIoU}$} \\ \hline
LVS$^*$ \cite{chen2021localizing}         &        \multirow{2}{*}{Flickr 10k}                      &       0.659                       &      0.529                    \\
\textbf{HTF (Ours)}   &                              & \textbf{0.718}                    & \textbf{0.558}                    \\ \hline
LVS$^*$ \cite{chen2021localizing}         &          \multirow{2}{*}{Flickr 144k}                    &       0.684                     &         0.535               \\
\textbf{HTF (Ours)}   &                              & \textbf{0.759}                    & \textbf{0.575}                    \\ \hline
LVS$^*$ \cite{chen2021localizing} & \multirow{2}{*}{VGGSound 144k} & 0.665 & 0.529 \\
\textbf{HTF (Ours)} & & \textbf{0.734} & \textbf{0.564} \\ \hline

\end{tabular}
\end{center}
\caption{Quantitative results on the novel, expanded Flickr SoundNet testing dataset where models are trained on the two training subsets of Flickr SoundNet and VGG Sound 144k. ``*" Denotes our faithful reproduction of the method.}
\label{tab:expandedflickrresult}
\end{table}

\section{Additional Implementation Details}

As mentioned in the main paper, we use ResNet18 feature extractors for the visual, audio, and flow portions of our method. For a given sample, the output features of the visual encoder, $f_v$, is a $7\times7$ spatial feature map where each spatial location has a feature vector of $512$ units. These features, once attended over with the optical flow localization network, are used with the audio representation of the sample to construct $S^{enh}$, the sound source localization map. During inference, $S^{enh}$ is upscaled to the size of the original image, which represents the visual sound source localization of that image. Furthermore, the attended visual and audio representations are both $L_2$ normalized before constructing $S^{enh}$.

For data augmentations, we randomly crop each image and optical flow map, in addition to a 50\% chance of applying a horizontal flip to both. We normalize the images using the standardized ImageNet normalization statistics and the optical flow maps using a mean of 0 and standard deviation of 1. 

\begin{table}[]
\begin{center}
\begin{tabular}{lccc}
\hline
Method       & Training Set                 & \multicolumn{1}{l}{cIoU$_{0.5}$} & \multicolumn{1}{l}{AUC$_{cIoU}$} \\ \hline
LVS \cite{chen2021localizing} & \multirow{2}{*}{Flickr 10k}  & 0.582 & 0.525 \\
LVS$^*$ \cite{chen2021localizing}         &                         &              0.730                 &      0.578               \\ \hline
LVS \cite{chen2021localizing} & \multirow{3}{*}{Flickr 144k}  & 0.699 & 0.573 \\
LVS$^\dagger$ \cite{chen2021localizing} & & 0.697 & 0.560 \\
LVS$^*$ \cite{chen2021localizing}      &   &           0.702                  &         0.588                   \\ \hline
LVS \cite{chen2021localizing} & \multirow{2}{*}{VGGSound 144k} & 0.719 & 0.582 \\
LVS$^*$ \cite{chen2021localizing} &  & 0.719 & 0.587 \\ \hline

\end{tabular}
\end{center}
\caption{Reproduction results of LVS \cite{chen2021localizing} on the Flickr SoundNet testing dataset. ``*" Denotes our faithful reproduction of the method, and ``$\dagger$" denotes our evaluation reproduction using officially provided model weights.}
\label{tab:flickrreprod}
\end{table}

\section{Additional Qualitative Results}

\begin{figure*}[t!]
\centering
\includegraphics[scale=0.16]{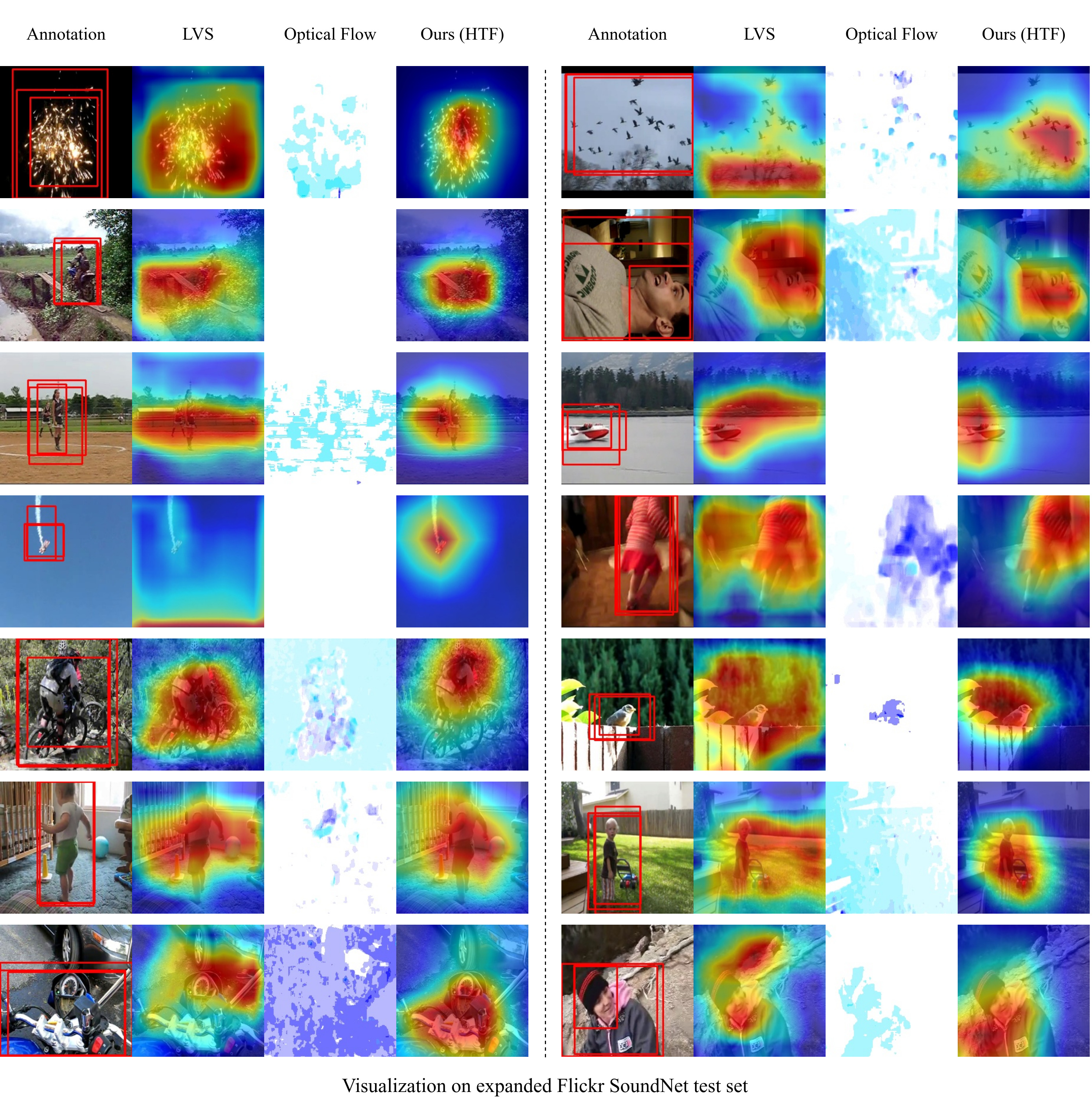}
\caption{Qualitative results of our method on the expanded Flickr SoundNet test set described in \ref{sec:expand}. Examples are from models trained on the Flickr 144k set. Figure is best viewed in color.}
\label{fig:qualbig}
\end{figure*}

In Figure \ref{fig:qualbig}, we visualize examples comparing our method against LVS \cite{chen2021localizing} on our expanded Flickr SoundNet test set, described in \ref{sec:expand}. As shown, our method is able to reliably localize towards the visual sound source, both in the presence and absence of meaningful optical flow information. Further, we show that these otherwise unused labeled samples are of high quality and are a useful addition for better evaluating self-supervised visual sound source localization methods. These annotated samples, previously reserved for training, are not needed for the self-supervised learning objective.

\section{Prior Work Reproduction}

As mentioned in \ref{dataconstruction}, since many of the videos in the dataset are missing, generating perfromance numbers for prior works on the available test videos becomes important to fairly assess the contribution of our work. We present a comparison of the performance of \cite{chen2021localizing} against our method, since it is the most relevant to our proposed approach. Reproducing other prior methods like \cite{song2022self} is a challenging task as the authors did not release pretrained models associated with their methods. Further, the public project repositories are missing relevant prepossessing or configuration files, which is required for a proper and fair reproduction of this method. To address this issue and further spur research, we open-source our code and other necessary resources for proper reproductions, available at \url{https://github.com/denfed/heartheflow}. 

As shown in Tables \ref{tab:flickrreprod} and \ref{tab:vggreprod}, we compare the reproduced results of LVS \cite{chen2021localizing} against the author's original results described in \cite{chen2021localizing}. For the Flickr SoundNet test set, our reproduced results are comparable with the original work. In certain cases, like training on Flickr 10k and Flickr 144k, we outperform the original results described in \cite{chen2021localizing}. For VGG Sound, the authors highlight that some VGG Sound Source annotations are updated and result in a 2-3\% difference in sound source localization performance on their official project page\footnote{\url{https://github.com/hche11/Localizing-Visual-Sounds-the-Hard-Way}}. This difference is consistent with our reproduced results. Based on these results, we believe that our reproduction is faithful and hence we are able to provide a fair comparison to our method.

\begin{table}[]
\begin{center}
\begin{tabular}{lccc}
\hline
Method       & Training Set                 & \multicolumn{1}{l}{cIoU$_{0.5}$} & \multicolumn{1}{l}{AUC$_{cIoU}$} \\ \hline
LVS \cite{chen2021localizing} & \multirow{2}{*}{VGGSound 10k}  &  - &  - \\
LVS$^*$ \cite{chen2021localizing}         &                         &              0.297              &      0.358               \\ \hline
LVS \cite{chen2021localizing} & \multirow{3}{*}{VGGSound 144k} & 0.344 & 0.382 \\
LVS$^\dagger$ \cite{chen2021localizing} & & 0.288 & 0.359 \\
LVS$^*$ \cite{chen2021localizing} &  & 0.301 & 0.361 \\ \hline

\end{tabular}
\end{center}
\caption{Reproduction results of LVS \cite{chen2021localizing} on the VGG Sound Source testing dataset. ``*" Denotes our faithful reproduction of the method, and ``$\dagger$" denotes our evaluation reproduction using officially provided model weights.}
\label{tab:vggreprod}
\vspace{-3mm}
\end{table}

\end{document}